\documentclass[conference]{IEEEtran}
\IEEEoverridecommandlockouts
\usepackage{array}
\usepackage{amssymb}
\usepackage{booktabs}
\usepackage{amsfonts} 
\usepackage[cmex10]{amsmath}
\usepackage{cite}
\usepackage{color,xcolor,soul,framed} 
\usepackage{eqparbox}

\usepackage{mdwmath}
\usepackage{mdwtab}
\usepackage{url}
\usepackage{multirow}

\usepackage{tikz}
\usepackage{tikz-qtree}
\usetikzlibrary{arrows,shapes,positioning,shadows,trees}

\usepackage{hyperref}
\hypersetup{
    colorlinks=true,
    citecolor=blue,
    linkcolor=red,
    filecolor=magenta,      
    urlcolor=cyan,
    pdftitle={Overleaf Example},
    pdfpagemode=FullScreen,    }

\begin{document}
\title{Information Aided Navigation: A Review}

\author{Daniel Engelsman and Itzik Klein, \IEEEmembership{Senior Member, IEEE}
	\thanks{Daniel Engelsman and Itzik Klein are with the Hatter Department of Marine Technologies, Charney School of Marine Sciences, University of Haifa, Israel. E-mails: \{dengelsm@campus, kitzik@univ\}.haifa.ac.il}}
\maketitle

\begin{abstract}
The performance of inertial navigation systems is largely dependent on the stable flow of external measurements and information to guarantee continuous filter updates and bind the inertial solution drift. Platforms in different operational environments may be prevented at some point from receiving external measurements, thus exposing their navigation solution to drift. Over the years, a wide variety of works have been proposed to overcome this shortcoming, by exploiting knowledge of the system current conditions and turning it into an applicable source of information to update the navigation filter. This paper aims to provide an extensive survey of information aided navigation, broadly classified into direct, indirect, and model aiding. Each approach is described by the notable works that implemented its concept, use cases, relevant state updates, and their corresponding measurement models. By matching the appropriate constraint to a given scenario, one will be able to improve the navigation solution accuracy, compensate for the lost information, and uncover certain internal states, that would otherwise remain unobservable.
\end{abstract}

\begin{IEEEkeywords}
Inertial Navigation Systems, Inertial Sensors, Extended Kalman Filter, Pseudo-Measurements, Non-Holonomic Constraints, Vehicle Constraints, Model-Based Navigation.
\end{IEEEkeywords}

\section{Introduction}
\begin{figure}[t]
\begin{center}
\includegraphics[width=0.53 \textwidth]{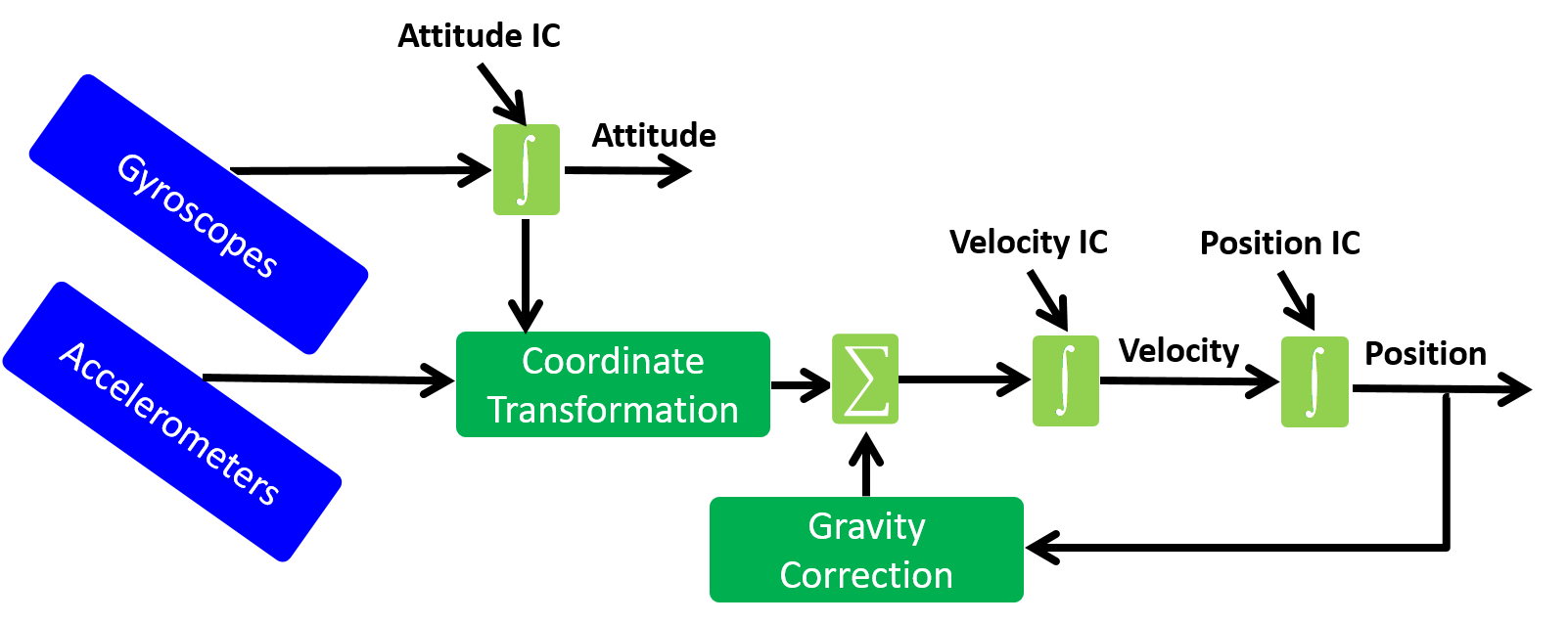}
\caption{Simplified block diagram of an inertial navigation system.}
\label{f:ins}
\end{center}
\end{figure}

\begin{figure*}[!t]
\begin{center}
\includegraphics[width=0.775 \textwidth]{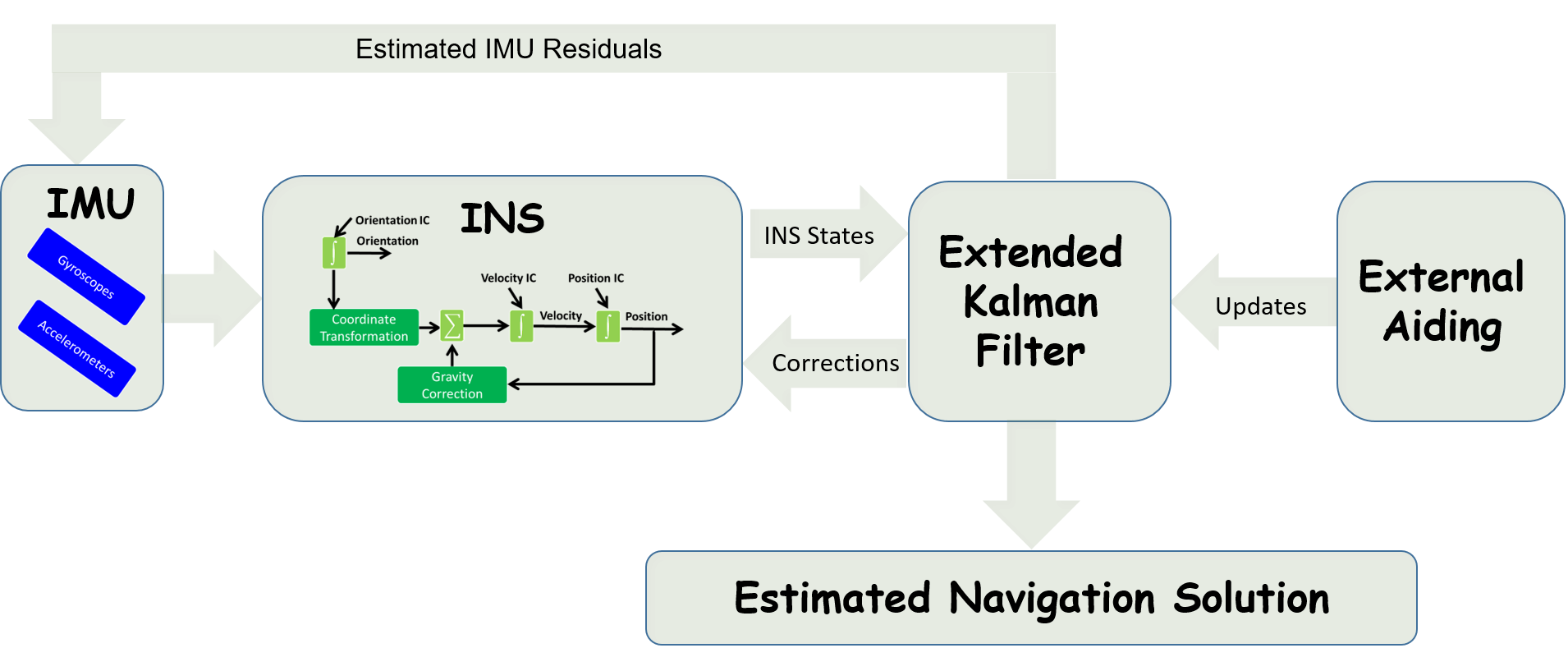}
\caption{Block diagram of the navigation filter with external measurements.}
\label{fig:ekf_fi}
\end{center}
\end{figure*}

\IEEEPARstart{I}{nertial} Navigation Systems (INS) are one of the most commonly used system for navigation. They can be found in almost every type of platform operating at land, sea, air, space, underground, and underwater  \cite{Titterton2004, Noureldin2013}. INS are also commonly used by humans for self-navigation \cite{may2003pedestrian} and are mounted on animals for research purposes \cite{nehmzow1995animal}. \\
INS contains inertial sensors, namely accelerometers and gyroscopes, capable of measuring the specific force and angular velocity vectors. INS uses these measurements to calculate position, velocity, and orientation at each epoch as shown in Figure~\ref{f:ins}. 
Its main advantages:
\begin{itemize}
	\item Provides a complete navigation solution for calculating position, velocity, and orientation.
	\item Operates in any environment, especially in challenging navigation environments such as indoors, underground, or underwater. 
	\item A standalone system, it does not rely on external broadcast or information to operate.
	\item Available in different grades, sizes, performance, and costs. 
\end{itemize}
The main disadvantage of the INS is its solution drift in situations of pure inertial operation \cite{Britting1971, Jekeli2000}. The inertial sensor measurements contain noises and other error terms that penetrate into the navigation solution during the integration process. Therefore, regardless of the INS quality, the navigation solution drifts over time. To mitigate that, it is fused with external sensors or information. The fusion process is carried out using a nonlinear filter, where commonly an extended Kalman filter (EKF) is applied \cite{Farrell2008, Groves2013}. During such a process, the external measurements are used to estimate both the navigation states and sensor error residuals as illustrated in Figure~\ref{fig:ekf_fi}.
The question if such a measurement can actually estimate the navigation states is addressed through observability analysis \cite{maybeck1982stochastic,klein2014observability}. There, vehicle dynamics and measurement type determine which states can be estimated \cite{li2012observability}. 
%
\tikzset{font=\small,
edge from parent fork down,
level distance=1.2cm,
every node/.style={
    bottom color=gray!9,
    top color=white!50,
    rectangle, 
    draw=black!125,          
    thick,
    align=center,
    text depth = 1pt,
    },
edge from parent/.style=
    {draw=black!150,        
    thick
    }}

\begin{figure*}[!hbt]
\centering
\begin{tikzpicture}
\tikzset{level 1/.style={level distance=30pt, sibling distance=4mm}}
\tikzset{level 2/.style={level distance=34pt, sibling distance=1.5mm}}
\tikzset{level 3+/.style={level distance=53pt, sibling distance=1.3mm}}

\Tree [.{\textbf{Information-aiding}}
        [.{\ref{sec_3}. Direct} 
            [.{Constant}
                [.{CA\\CP\\CHA} ]
            ]
            [.{Zero} 
                [.{ZVN\\ZVB} ]
                [.{ZAR\\ZYR} ]
                [.{ZDV\\ZAD\\ZAN} ]
            ]
            [.{Relative} 
                [.{VLA\\RUP} ]
            ]
        ]
        [.{\ref{sec_4}. Indirect}
            [.{Hetero.}
                [.{PbO\\VbO} ]
            ]
            [.{Homog.}
                [.{P-GPS\\B-DVL} ]
            ]
        ] 
        [.{\ref{sec_5}. Model-aided}
            [.{Aerial} 
                [.{ADM\\MAIN\\FVM} ]
            ]
            [.{Marine}
                [.{ASC\\CFD\\POM} ]
            ]
            [.{Underwater}
                [.{AUV\\NLDM} ]
            ]
            [.{Land}
                [.{VKM\\NHC} ]
            ]
        ]
]
\end{tikzpicture}
\caption{Taxonomy tree of information aided navigation.} \label{f:Map-Taxonomy}
\end{figure*}
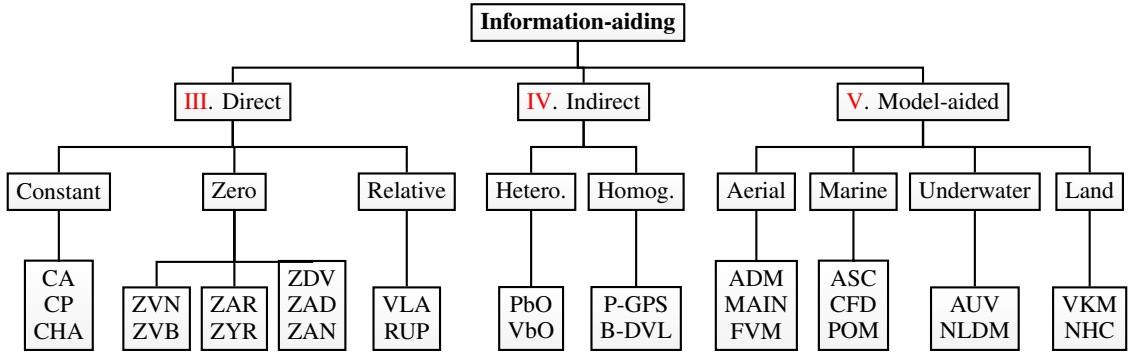

Focusing on the aiding type, a distinction is made between external sensor aiding and information aiding. External sensor aiding refers to the physical sensor itself, which provides external measurements to aid the INS. Radio frequency (RF) navigation, for example, offers GNSS broadcasting for outdoors \cite{wendel2006integrated, hasan2009review, toth2016remote}, RF-based technologies \cite{kubitz1997application, raquet2008non, moghtadaiee2011indoor} and Wi-Fi \cite{biswas2010wifi, kotaru2015spotfi, vasisht2016decimeter} for indoors. Given a GNSS-denied environment, acoustic navigation can be used, especially underwater, using sonar \cite{1087534, elfes1987sonar, leonard2012directed}, range-based localization techniques \cite{mcphail2009range, bayat2013auv, bayat2015range}, long and short baseline systems \cite{smith1997experimental, collin2000spatial, larsen2000synthetic, jakuba2008long}, and a Doppler velocity log (DVL) \cite{snyder2010doppler, gao2014strapdown, chang2016initial}.\\
A different type of external sensing is a geographical navigation, where the platform's surroundings are exploited for measurements, e.g., star trackers \cite{clark2000intelligent, brown2007gps, rad2014optimal, capuano2014gnss}, celestial navigation \cite{van2004basic, ning2007autonomous, karl2007celestial}, Earth's magnetic field \cite{caruso2000applications, yang2003magnetometer, goldenberg2006geomagnetic}, and recently vision-aided navigation, where visual landmarks are used for the navigation solution \cite{gaspar2000vision, sinopoli2001vision, madison2007vision, caballero2009vision}. 
In contrast to external sensors, information aiding does not require a physical sensor and may be used to update the navigation states with and without external sensors. Information aiding has two main advantages over external sensor aiding:
\begin{itemize} 
\item \textbf{Cost free}: Information aiding does not involve any actual sensors, but only a once-off software modification.
\item \textbf{Continuous availability}: The update rate  can be conveniently set to the INS operating sampling rate or any other desired rate.
\end{itemize}
In this paper, we aim to highlight the importance of information aiding navigation and provide a state-of-the-art review of the different aiding approaches, operating scenarios, assumptions, and references for further reading. To that end, we categorize information aiding into groups:
\begin{itemize}
    \item \textbf{Direct information aiding}: Knowledge about the INS carrying platform or its working environment can be translated directly into information measurements, aiding the navigation filter.
    \item \textbf{Indirect information aiding}: In some scenarios additional information can be extracted from external sensors and imposed indirectly to aid the navigation filter.  
    \item \textbf{Model-based aiding}: A vehicle simulation runs in parallel to the INS model, enabling simultaneous fusion with the navigation solution, such that both feed each other reciprocally and enable state estimation of their errors. 
\end{itemize}
Each of those information aiding categories is thoroughly addressed in this paper. A taxonomy tree of all 25 different information aiding types is presented in Figure~\ref{f:Map-Taxonomy}. 
The rest of the paper is organized as follows: Section \ref{sec_3} introduces direct aiding techniques, Section \ref{sec_4} describes indirect aiding powered by external sensors. Section \ref{sec_5} presents model-aided approaches, and Section \ref{sec_conc} gives the conclusions. 

\section{Navigation Filter}\label{sec:NavFilter}
Fusion of INS with external sensors or data requires a nonlinear filter due to the nonlinear nature of the INS equations of motion. Mostly, an error-state EKF implementation is used with a state vector $\delta \mathbf{x} = \Tilde{\mathbf{x}} - \mathbf{x} \in \mathbb{R}^{15}$. It defines the difference between the true state vector and the filter estimate, thus yielding the following error terms:
\begin{align}
\Tilde{\textbf{p}}^n &= \textbf{p}^n + \delta \textbf{p}^n \label{err_p} \\
\Tilde{\textbf{v}}^n &= \textbf{v}^n + \delta \textbf{v}^n \\
\Tilde{\textbf{T}}_b^n &= (\textbf{I} + [\boldsymbol{\epsilon}^n \times] ) \textbf{T}_b^n \\
\Tilde{\boldsymbol{f}}_{ib}^b &= \boldsymbol{f}_{ib}^b + \textbf{b}_a + \textbf{w}_a \\
\Tilde{\boldsymbol{\omega}}_{ib}^b &= \boldsymbol{\omega}_{ib}^b + \textbf{b}_g + \textbf{w}_g \label{err_b_g}
\end{align}
$\textbf{I}$ denotes a $3 \times 3$ identity matrix, $[\times]$ is a skew-symmetric form of a vector, and $\textbf{T}_b^n$ is the transformation matrix from body to navigation frame. The position error vector $\delta\mathbf{p}^{n}$, velocity error vector $\delta\mathbf{v}^{n}$, misalignment angles errors $ \boldsymbol{\epsilon}^{n}$, accelerometer bias residuals $\delta \boldsymbol{f}_{ib}^b \approx \mathbf{b}_{a}$, and gyro bias residuals $\delta \boldsymbol{\omega}_{ib}^b \approx \mathbf{b}_{g}$ \cite{Noureldin2013, Farrell2008, Groves2013}, together make up the error state vector:
\begin{equation}\label{eq_iInsErrorState}
\delta \mathbf{x} = \left[ \begin{array}{ccccc} 
\delta\mathbf{p}^{n} & \delta\mathbf{v}^{n} & \mathbf{\epsilon}^{n} & \mathbf{b}_{a} & \mathbf{b}_{g} \end{array} \right]^{\operatorname{T}}
\end{equation}
where the accelerometers and gyro biases are expressed in the body frame and the superscript $n$ denotes a vector expressed in the navigation frame.
The linearized error state model is \cite{Groves2013}
\begin{equation}\label{eq_errorModel}
\delta\dot{\mathbf{x}} = \mathbf{F}\delta\mathbf{x}  + \mathbf{G}\delta\mathbf{w}
\end{equation}
where $\mathbf{F}$ is the system matrix, $\mathbf{G}$ is the shaping matrix, and $\delta\mathbf{w}$ is the noise vector.
The accelerometers and gyros residuals are modeled as random walk process although any other suitable models could be used instead such as the first-order Gauss-Markov processes. The system matrix is given by
\begin{equation}\label{eq_Fmat}
\mathbf{F} = \left[ \begin{array}{ccccc}
\mathbf{F}_{pp} & \mathbf{F}_{pv} & \mathbf{0}_{3 \times 3} & \mathbf{0}_{3\times 3} & \mathbf{0}_{3\times 3} \\
\mathbf{F}_{vp} & \mathbf{F}_{vv} & \mathbf{F}_{v\epsilon} & \mathbf{T}_{b}^{n} & \mathbf{0}_{3\times 3} \\
\mathbf{F}_{\epsilon p} & \mathbf{F}_{\epsilon v} & \mathbf{F}_{\epsilon\epsilon} & \mathbf{0}_{3\times 3} & \mathbf{T}_{b}^{n} \\
\mathbf{0}_{3\times3} & \mathbf{0}_{3\times 3} & \mathbf{0}_{3\times 3} & \mathbf{0}_{3\times 3} & \mathbf{0}_{3\times 3} \\
\mathbf{0}_{3\times3} & \mathbf{0}_{3\times 3} & \mathbf{0}_{3\times 3} & \mathbf{0}_{3\times 3} & \mathbf{0}_{3\times 3} 
\end{array}
\right]. 
\end{equation}
$F_{ij}$ is a $3\times 3$ submatrix as a result of the linearization of the nonlinear equation of motion (more details on the internalization process can be found in navigation textbooks such as \cite{Farrell2008, Groves2013}). The shaping matrix is given by
\begin{equation}\label{eq_Gmat}
\mathbf{G} = \left[ \begin{array}{cccc}
\mathbf{0}_{3\times3} & \mathbf{0}_{3\times 3} & \mathbf{0}_{3\times 3} & \mathbf{0}_{3\times 3} \\
\mathbf{T}_{b}^{n} & \mathbf{0}_{3\times 3} & \mathbf{0}_{3\times 3} & \mathbf{0}_{3\times 3} \\
\mathbf{0}_{3\times3} & \mathbf{T}_{b}^{n} & \mathbf{0}_{3\times 3} & \mathbf{0}_{3\times 3} \\
\mathbf{0}_{3\times3} & \mathbf{0}_{3\times 3} & \mathbf{I}_{3} & \mathbf{0}_{3\times 3} \\ 
\mathbf{0}_{3\times3} & \mathbf{0}_{3\times 3} &  \mathbf{0}_{3\times 3} &\mathbf{I}_{3} 
\end{array}
\right] 
\end{equation}
and the noise vector is
\begin{equation}\label{eq_NoiseVec}
\delta \mathbf{w} = \left[ \begin{array}{cccc} 
\mathbf{w}_a & \mathbf{w}_g & \mathbf{w}_{ab} & \mathbf{w}_{ab} \end{array} \right]^{\operatorname{T}}
\end{equation}
where $\mathbf{w}_a$ and $\mathbf{w}_g$ are the process residual noises and $\mathbf{w}_{ab}$ and $ \mathbf{w}_{gb}$ are the sensor measurement noises of the accelerometers (a) and gyroscopes (g), respectively. All noises are modelled as zero-mean white Gaussian distribution, assumed to be constant for all samples. Following \cite{Farrell2008}, the EKF error-state closed loop implementation algorithm is
\begin{eqnarray}\label{eq_EKF}
\delta\hat{\mathbf{x}}^{-}_{k} & = & 0 \\
\mathbf{P}^{-}_{k} & = & \mathbf{\Phi}_{k-1}\mathbf{P}^{+}_{k-1}\mathbf{\Phi}^{\operatorname{T}}_{k-1}+\mathbf{Q}_{k-1}\\
\delta\hat{\mathbf{x}}^{+}_{k} & = & \mathbf{K}_{k}\delta\mathbf{z}_{k} \\
\mathbf{P}^{+}_{k} & = & [\mathbf{I}-\mathbf{K}_{k}\mathbf{H}_{k}]\mathbf{P}^{-}_{k} \\
\mathbf{K}_{k} & = & \mathbf{P}^{-}_{k}\mathbf{H}^{\operatorname{T}}_{k}[\mathbf{H}_{k}\mathbf{P}^{-}_{k}\mathbf{H}^{\operatorname{T}}_{k}+\mathbf{R}_{k}]^{-1} \label{eq_EKFend}
\end{eqnarray}
where $k$ is the time step index, $\delta\mathbf{x}^{-}_{k}$ is the a priori estimate of the error-state, $\delta\mathbf{x}^{+}_{k}$ is the \textit{a posteriori} estimate of the error-state, $\delta \mathbf{z}_{k}$ is the measurement residual vector, $\mathbf{P}^{-}_{k}$ is the covariance of the a priori estimation error, $\mathbf{P}^{+}_{k}$ is the covariance of the \textit{a posteriori} estimation error and $\mathbf{K}_{k}$ is the Kalman gain. $\mathbf{Q}_{k}$ is the process noise covariance and $\mathbf{R}_{k}$ is the measurement noise covariance, both assumed to be constant for all samples, and $\mathbf{\Phi}_{k}$ is the state transition matrix.
The measurement matrix $\mathbf{H}_{k}$ contains basis vectors that adjust the relevant INS states to the information aided update $\mathbf{z}_k$ and $\boldsymbol{\nu}_{\textbf{i}}$ denotes the mean white Gaussian measurement noise vector of the i-th external aiding type, where non-bold ${\nu}_{\text{i}}$ denotes scalar.
\begin{figure}[!h]
\centering 
\includegraphics[width=.485\textwidth, clip, keepaspectratio]{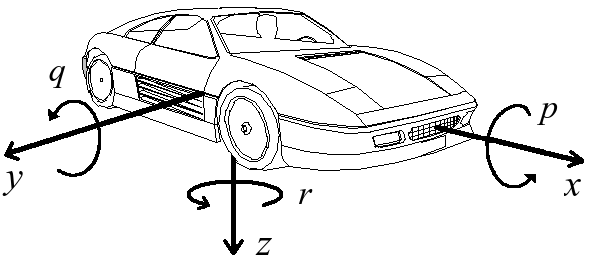}
\caption{Body frame coordinate system.}
\label{fig:bodyframe}
\end{figure}\\
Figure~\ref{fig:bodyframe} illustrates the body frame coordinate system, showing the $x\text{-}y\text{-}z$ axes definitions with the corresponding $p\text{-}q\text{-}r$ angular velocities around each of them. The navigation frame center is located at the body center of mass, where the $x$-axis points to the geodetic north, the $z$-axis points down parallel to the local vertical, and the $y$-axis completes a right-handed orthogonal frame. The body frame center is located at the center of mass, where the $x$-axis is parallel to the longitudinal axis of symmetry of the vehicle pointing forward, the $y$-axis points right, and the z-axis points down such that it forms a right-handed orthogonal frame. 

\section{Direct Information Aiding} \label{sec_3} 
\begin{figure}[!h]
\begin{center}
\includegraphics[width=0.48 \textwidth]{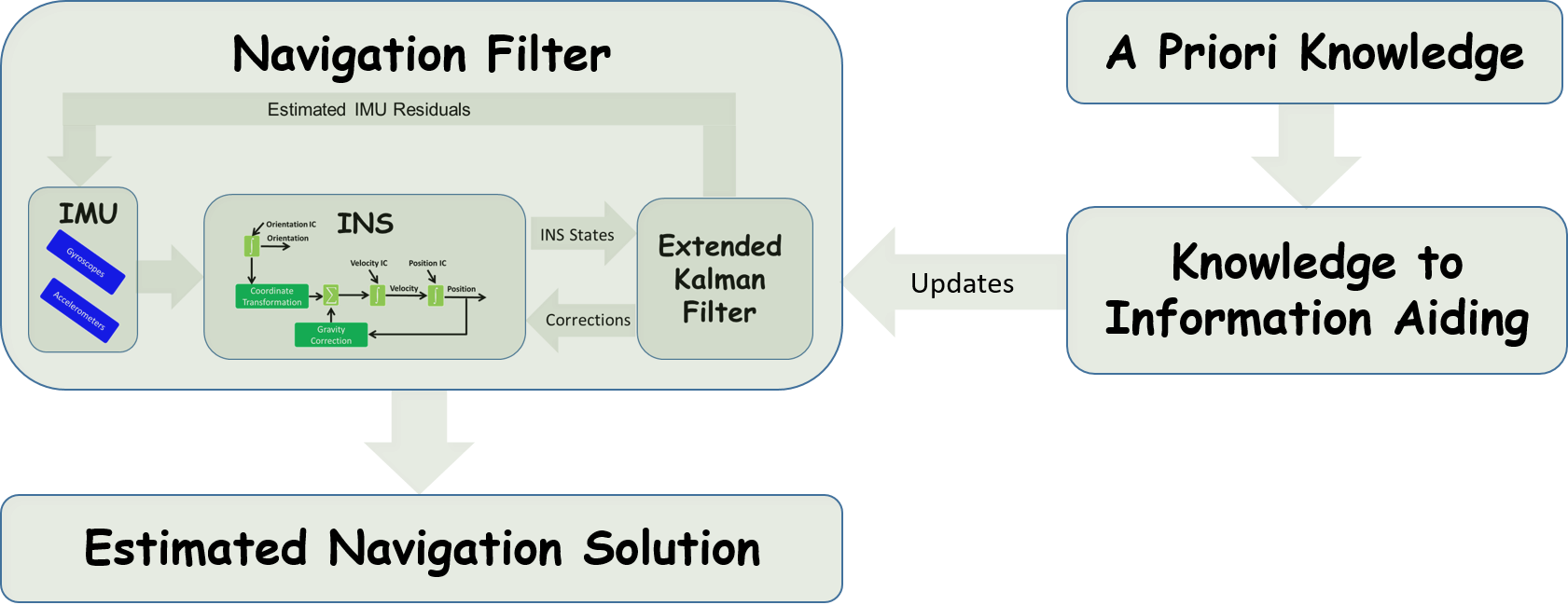}
\caption{Block diagram of the navigation filter with information aiding.}
\label{fig:ekf_ia}
\end{center}
\end{figure}

\noindent
This section surveys a wide variety of techniques that use a priori information about the platform dynamics, translate it into external measurements and feed it into the navigation filter. For each information measurement in this section, the underlying assumption of the measurement is stated, followed by its applications, working principle, measurement model, and examples from the literature.\\
Table~\ref{tab:IA} summarizes all 12 aiding types and their abbreviations, and gives hyperlinks to their location in this section.
\begin{table}[!h] 
\centering
\caption{Direct aiding types} \label{tab:IA}
\begin{tabular}{ >{\centering}p{0.8cm} >{\centering}p{1cm}  p{4.5cm} }
 Section & Abbrv. & \hspace{10mm} Aiding Type \\ [2mm] \hline \\
\ref{zero_velocity_nav} & ZVN & Zero velocity in navigation frame  \\ [1.40mm]
\ref{zero_velocity_body} & ZVB & Zero velocity in body frame  \\ [1.40mm]
\ref{zero_angular_rate} & ZAR & Zero angular rates in body frame  \\ [1.40mm]
\ref{const_height} & CA & Constant altitude \\ [1.40mm]
\ref{const_down_velocity} & ZDV & Zero down velocity \\ [1.40mm] 
\ref{zero_acc_down} & ZAD & Zero acceleration (down) in body frame \\ [1.40mm]
\ref{zero_acc_navigation} & ZAN & Zero acceleration in navigation frame \\ [1.40mm]
\ref{const_position} & CP & Constant position \\ [1.40mm]
\ref{zero_yaw_rate} & ZYR & Zero yaw rate \\ [1.40mm]
\ref{const_heading_angle} & CHA & Constant heading angle \\ [1.40mm]
\ref{lever_arm} & VLA & Virtual lever arm \\ [1.40mm]
\ref{m_imus} & RUP & Relative update \\ [1.40mm]
\hline \end{tabular} 
\end{table}\\
Figure~\ref{fig:ekf_ia} illustrates how the direct information approach acts effectively as an aiding mean in the navigation filter, thus substituting the external sensor.
%
\subsection{Zero Velocity in Navigation Frame (ZVN)} \label{zero_velocity_nav}
\noindent
\textbf{Assumption}: Stationary conditions. \\
\textbf{Applications}: Land vehicles, mobile robots, shoe-mounted pedestrian navigation, and stationary fine alignment. \\
\textbf{Principle}: While stationary, the velocity vector expressed in navigation frame $\textbf{v}^n$ is assumed to be zero. \\
\textbf{Measurement Model}: The measurement residual is given by:
\begin{align}
\delta \mathbf{z}_{_{\textbf{ZVN}}} &= \textbf{v}_{_{\text{INS}}}^n - \textbf{0}_{3 \times 1} = \textbf{H}_{_{\textbf{ZVN}}} \delta \mathbf{x} + \boldsymbol{\nu}_{_{\textbf{ZVN}}}
\end{align}
where $\textbf{v}_{_{\text{INS}}}^n$ is the calculated INS velocity vector and $\boldsymbol{\nu}_{_{\textbf{ZVN}}}$ is zero mean white Gaussian measurement noise. The corresponding measurement matrix is defined by:
\begin{equation}
\textbf{H}_{_{\textbf{ZVN}}} = \begin{bmatrix}
\textbf{0}_{3 \times 3} & \textbf{I}_{3 \times 3} & \textbf{0}_{3 \times 3} & \textbf{0}_{3 \times 3} & \textbf{0}_{3 \times 3}
\end{bmatrix}.
\end{equation} 
\textbf{In the Literature}: For land vehicles, zero velocity updates, also known as ZUPT, were applied by Salychev \cite{salychev1998inertial} on vehicles when approaching a stop sign, or when standing at a red light. ZVU used were also to bridge situations of  global positioning system (GPS) outages in urban canyons \cite{grejner2001bridging, grejner2002bridging}. Xiaofang \cite{xiaofang2014applications} examined its contribution over land vehicle simulation using three kinds of zero-velocity detectors and Ben \cite{ben2009improved} presented ZUPT-based estimation filter using a separate-bias Kalman filter. When using low-cost sensors, ZVN can be also useful to improve attitude accuracy once zero velocity is detected \cite{suh2011smoother, duong2015computationally, saadeddin2014estimating, duong2016simple, klein2021ins}. \\
In shoe-mounted INS, user stationary stance periods are identified followed by a ZVN update to the navigation filter \cite{foxlin2005pedestrian, skog2010zero, skog2010evaluation, fischer2012tutorial, rantakokko2014foot, zhuang2015pdr, ruppelt2016high, song2018enhanced}. Fourati \cite{fourati2014heterogeneous} proposed heterogeneous data fusion techniques for pedestrian navigation and Tong \cite{tong2019double} proposed Hidden Markov model (HMM)-based ZVN for pedestrian dead reckoning navigation. Other works suggested ZVN updates based on parametric thresholds over the gait cycle \cite{faulkner2010altitude, li2012robust, abdulrahim2014understanding, ren2016novel, zhang2017adaptive}, aiding tightly coupled integration \cite{sun2020bds} for an observability analysis of the strapdown INS alignment \cite{wu2012observability}, and an open-source foot-mounted module combined with embedded software for pedestrian applications \cite{nilsson2012foot, nilsson2014foot}.
ZVN are also used in the stationary FA process. Bar-Itzhack \cite{bar1988control} presented a control theory point of view for a linear error model, enabling to determine the INS unobservable subspace, and recently Silva \cite{silva2018measurement} used it to aid stationary FA. Suh \cite{suh2013distance} incorporated ZVN with vision-aided navigation, Noureldin \cite{noureldin2002accuracy} used it for real-time positioning of horizontal measurement-while-drilling (MWD) in the oil industry, and Li \cite{li2012dead} extended its definition into a constant velocity update (CUPT), which handles stance phases whenever velocity cannot be assumed as zero, e.g., while standing in a moving elevator or escalators.

\subsection{Zero Velocity in Body Frame (ZVB)} \label{zero_velocity_body}
\noindent
\textbf{Assumption}: Motion on a planar surface without slipping. \\ 
\textbf{Applications}: Land vehicles and mobile robots. \\ 
\textbf{Principle}: In land vehicles, given no-slip conditions, the velocity components perpendicular to the forward direction, i.e., body $y$-axis (transverse) and body  $z$-axis (down), (see Figure~\ref{fig:bodyframe} for axes definition) are assumed as zero:
\begin{align} \label{eq_non_holo}
\textbf{v}^b_{yz} = \begin{bmatrix} v^b_y \\ v^b_z \end{bmatrix} &= \begin{bmatrix} \textbf{e}^{\operatorname{T}}_2 \\ \textbf{e}^{\operatorname{T}}_3 \end{bmatrix} \textbf{v}^b = \textbf{0}_{2 \times 1}
\end{align}
where $\textbf{e}_i$ for $1\leq i \leq 3$ is a basis for $\mathbb{R}^{3}$ and $\textbf{v}^b$ is the velocity vector expressed in the body frame.\\
\textbf{Measurement Model}: The measurement residual is given by:
\begin{align}\label{eq:zvb1}
\delta \mathbf{z}_{_{\textbf{ZVB}}} &= \textbf{v}^b_{\text{INS},yz} - \textbf{0}_{2 \times 1}  = \textbf{H}_{_{\textbf{ZVB}}} \delta \mathbf{x} + \boldsymbol{\nu}_{_{\textbf{ZVB}}}
\end{align}
where $ \textbf{v}^b_{\text{INS},yz} $ is the calculated INS body velocity, $\boldsymbol{\nu}_{_{\textbf{ZVB}}}$ is zero mean white Gaussian measurement noise, and $\textbf{H}_{_{\textbf{ZVB}}}$ is the corresponding measurement matrix. \\
As the body velocity vector is a function of the body to navigation transformation matrix, small error perturbation analysis is applied on (\ref{eq:zvb1}) to obtain the measurement matrix:
\begin{align}\label{eq:zvb2}
\delta \textbf{v}^b &= \Tilde{\textbf{v}}^b - \textbf{v}^b = (\Tilde{\textbf{T}}_{n}^b \Tilde{\textbf{v}}^n) - \textbf{v}^b \notag \\
&= \Big( \textbf{T}_n^b (\textbf{I} - [ \boldsymbol{\epsilon}^n \times ] )  \textbf{v}^n + \textbf{T}_n^b \delta \textbf{v}^n \Big) - \textbf{v}^b \notag \\
&= \textbf{T}_n^b \delta \textbf{v}^n - \textbf{T}_n^b [ \textbf{v}^n \times ] \boldsymbol{\epsilon}^n .
\end{align}
Finally, from (\ref{eq:zvb2}), the measurement matrix is defined by: 
\begin{equation}
\textbf{H}_{_{\textbf{ZVB}}} = 
\begin{bmatrix} \textbf{e}^{\operatorname{T}}_2 \\ \textbf{e}^{\operatorname{T}}_3 \end{bmatrix} \begin{bmatrix}
\textbf{0}_{3 \times 3} & \textbf{T}_n^b & -\textbf{T}_n^b [\textbf{v}^n \times] & \textbf{0}_{3 \times 3} & \textbf{0}_{3 \times 3}
\end{bmatrix}.
\end{equation}

\textbf{In the Literature}: Zero velocity in the body frame dates back to the 2000's, where Brandt \cite{Brandit1998}, Sukkarieh \cite{sukkarieh2000low}, Collin \cite{collin2001unaided}, and later Dissanayake \cite{dissanayake2001aiding} exploited it as vehicle model constraints. Shin \cite{shin2002accuracy}, Liu \cite{liu2012performance}, and later Falco \cite{falco2017loose, falco2017benefits} showed how ZVB maintains the INS errors bounded in case of GPS outage, using tightly coupled (TC) GNSS/INS architecture. Godha (2005) \cite{godha2005integration} and later Chiang \cite{chiang2013performance} focused on its benefits for long-term GPS outages in the open, and Syed (2008) \cite{syed2008civilian} used it to examine its effects on misalignment accuracy. Later, analytical observability analysis determined which states are unobservable in the navigation filter using ZVB\cite{elsheikh2018low}. Other methods incorporated odometry or wheel speed sensors (WSS) along with the ZVB constraint to update the body velocity  vector \cite{niu2005development, gao2006development, gao2006gps, niu2007accurate, georgy2009low, li2010ultra, li2011real, li2009performance, wang2010land, chen2018sins, chen2019low, zhu2020mimu}, including for trains \cite{chen2013railway, reimer2016ins, chen2018railway, zhou2019kinematic, zhang2019requirement, sun2021high}, precision agriculture \cite{yang2014enhanced, lan2019integrated, zhang2020velocity,  zhang2021evaluating}, indoor localization \cite{won2015performance}, and a Map/INS/Wi-Fi integrated system \cite{yu2017map}.

\subsection{Zero Angular Rate in Body Frame (ZAR)} \label{zero_angular_rate}
\noindent
\textbf{Assumption}: Stationary conditions. \\
\textbf{Applications}: Land vehicles, mobile robots, shoe-mounted navigation, and stationary fine alignment. \\
\textbf{Principle}: While stationary, the angular rate vector between the navigation and ECEF frames, $\boldsymbol{{\omega}}_{en}^n$, is zero and for low-cost sensors, the Earth rotation rate, $\boldsymbol{{\omega}}_{ie}^n$, can be neglected. Thus, if the gyros output, $\boldsymbol{{\omega}}_{ib}^b$, is approximately zero, the ZAR update is given by
\begin{equation}
\boldsymbol{\omega}^b_{nb} = \boldsymbol{\omega}_{ib}^b - (\boldsymbol{\omega}_{ie}^b + \boldsymbol{\omega}_{en}^b ) = \textbf{0}_{3 \times 1}.
\end{equation}
\textbf{Measurement Model}: The measurement residual is given by:
\begin{equation} \label{eq_delta_w}
\delta \mathbf{z}_{_{\textbf{ZAR}}} = \boldsymbol{{\omega}}^b_{nb,_{{\text{INS}}}} - \textbf{0}_{3 \times 1} = \textbf{H}_{_{\textbf{ZAR}}} \delta \mathbf{x} + \boldsymbol{\nu}_{_{\textbf{ZAR}}} = \textbf{b}_g+ \boldsymbol{\nu}_{_{\textbf{ZAR}}}
\end{equation}
where $\boldsymbol{{\omega}}^b_{nb,_{{\text{INS}}}}$ is the calculated INS body angular rate vector, $\boldsymbol{\nu}_{_{\textbf{ZAR}}}$ is zero mean white Gaussian measurement noise, $\textbf{b}_g$ is the gyroscope bias vector expressed in the body frame, and the corresponding measurement matrix is defined by:
\begin{equation}
\textbf{H}_{_{\textbf{ZAR}}} = \begin{bmatrix}
\textbf{0}_{3 \times 3} & \textbf{0}_{3 \times 3} & \textbf{0}_{3 \times 3} & \textbf{0}_{3 \times 3} & \textbf{I}_{3 \times 3}
\end{bmatrix}.
\end{equation} 
\textbf{In the Literature}: 
Zero angular rate update, also known as ZARU, was implemented by Ramanandan for observability analysis and tracking the error states of an INS aided with stationary updates \cite{ramanandan2011observability, ramanandan2011inertial}. Groves \cite{groves2015navigation} and later Klein \cite{klein2021ins} applied ZAR to improve the stationary fine alignment process using low-cost sensors. Chen \cite{chen2012low} and Brossard \cite{brossard2019rins} used it for self-localizing indoor mobile robots, and Chow incorporated it in a joint sensor self-calibration method \cite{chow2018tightly}. \\
ZAR can be also used in navigation filters for pedestrian dead reckoning (PDR) applications, as seen by Jiménez \cite{jimenez2010indoor}, Bancroft \cite{bancroft2012estimating}, and later by Benzerrouk \cite{benzerrouk2014mems, benzerrouk2018robust}. Woyano \cite{woyano2016evaluation} used it to improve EKF performance, and Zhang \cite{zhang2017standing, zhang2017foot} incorporated ZAR with a standing calibration method to improve calculations of human motion angular velocities, whereas Bar-Shalom \cite{bar2011tracking} noted that it should be approached with caution, as the residual angular rate of a stationary person is significantly larger than that of a stationary vehicle. \\
Other works include the ZAR as counterpart aiding during long-term GPS outages in urban environments \cite{klein2010pseudo, klein2011vehicle}, \\
aiding Wi-Fi based indoor localization \cite{chai2012ins, chai2012enhanced, chai2019multi}, and fusion with an error-based KF for INS/DVL applications \cite{karmozdi2018design}.
%
\subsection{Constant Altitude (CA)} \label{const_height} 
\noindent
\textbf{Assumption}: The altitude remains fixed during operation. \\
\textbf{Application}: Land vehicles, mobile robots, indoor navigation, and ships. \\
\textbf{Principle}: While moving on levelled surfaces, altitude remains constant, $h_c$, for short time intervals or for the entire trajectory.\\ 
\textbf{Measurement Model}: The measurement residual is given by:
\begin{equation}
\delta \mathbf{z}_{_{\textbf{CA} }} = h_{_{\text{INS}}} - h_c = \textbf{H}_{_{\textbf{CA} }} \delta \mathbf{x} + {\nu}_{_{\textbf{CA} }}
\end{equation}
where $h_{_{\text{INS}}}$ is the calculated INS altitude, ${\nu}_{_{\textbf{CA}}}$ is zero mean white Gaussian measurement noise, and the measurement matrix is defined by:
\begin{align}
\textbf{H}_{_{ \textbf{CA} }} = \begin{bmatrix}
\, \textbf{e}^{\operatorname{T}}_3 & \textbf{0}_{1 \times 3} & \textbf{0}_{1 \times 3} & \textbf{0}_{1 \times 3} & \textbf{0}_{1 \times 3}
\end{bmatrix}.
\end{align} 
\textbf{In the Literature}: Constant altitude update is used in vehicular urban navigation, where GPS signals are often obstructed by the surrounding buildings \cite{zhao2011gps, kaplan2017understanding, klein2010pseudo}. Godha \cite{godha2007gps} and later Angrisano added pseudo-height observations to improve their proposed GNSS-derived heading algorithm using low-cost MEMS IMUs \cite{angrisano2010gnss, angrisano2012benefits}.\\
In indoor pedestrian navigation systems, CA exploits the fact that an indoor floor is usually flat, and height barely changes unless using stairs, escalators, or elevators \cite{stirling2005evaluation, abdulrahim2012using, groves2015navigation, norrdine2016step, yoon2016robust, munoz2018height}. Munoz \cite{munoz2018height} and later Refai \cite{refai2020portable} applied CA based on biomechanical constraints related to the gait pattern, and Weenk \cite{weenk2014ambulatory} used CA to perform an ambulatory estimation of relative foot positions. \\
In marine applications, Ryan \cite{ryan1998augmentation} showed the advantage of a combined GPS-GLONASS system with height constraint, given calm sea conditions.
\subsection{Zero Down Velocity (ZDV)} \label{const_down_velocity} \noindent
\textbf{Assumption}: The platform down velocity component is assumed as zero during operation. \\
\textbf{Application}: Land vehicles, mobile robots, indoor navigation, and ships. \\
\textbf{Principle}: In addition to the constant height constraint in Section~(\ref{const_height}), motion on levelled surfaces also imposes zero vertical velocity constraint, for short time intervals or for the entire trajectory.\\
\textbf{Measurement Model}: The measurement residual is given by:
\begin{equation}
\delta z_{_{\textbf{ZDV}}} = v_{D_{\text{INS}}} - 0 = \textbf{H}_{_{\textbf{ZDV}}} \delta \mathbf{x} + {\nu}_{_{\textbf{ZDV}}}
\end{equation}
where $v_{D_{\text{INS}}}$ is the calculated INS vertical velocity component, ${\nu}_{_{\textbf{ZDV}}}$ is zero mean white Gaussian measurement noise, and the measurement matrix is defined by:
\begin{align}
\textbf{H}_{_{ \textbf{ZDV} }} = 
\begin{bmatrix}
\,  \textbf{0}_{1 \times 3} & \textbf{e}^{\operatorname{T}}_3 & \textbf{0}_{1 \times 3} & \textbf{0}_{1 \times 3} & \textbf{0}_{1 \times 3}
\end{bmatrix}.
\end{align} 
\textbf{In the Literature}: Zero down velocity is used to enhance vehicular INS navigation in urban environments \cite{klein2011vehicle}, using tightly coupled integration \cite{falco2017benefits} or for improving Wi-Fi based indoor vehicle navigation \cite{chai2012ins, chai2012enhanced}. It is also implemented in an odometry/gyro UKF for robot localization \cite{chen2012low} and in INS/DVL integrated navigation in shallow water \cite{karmozdi2018design}. 
\subsection{Zero Down Acceleration in Body Frame (ZAD)} \label{zero_acc_down} 
\noindent
\textbf{Assumption}: Motion on a planar surface. \\
\textbf{Applications}: Land vehicles and mobile robots. \\
\textbf{Principle}: Travelling at low velocities with low cost sensors allows neglecting the earth turn rate and the transport rate. Consequently, for short time periods, if the altitude and down velocity components are constant (CA and ZDV), the down acceleration in the body frame can be assumed to be zero.\\ 
\textbf{Measurement Model}: The measurement residual is given by: 
\begin{equation}
\delta z_{_{\textbf{ZAD}}} = a^n_{z,_{\text{INS}}} - 0 = b_{a_z} = \textbf{H}_{_{\textbf{ZAD}}} \delta \mathbf{x} + \nu_{_{\textbf{ZAD}}}
\end{equation}
where $a^n_{z,_{\text{INS}}}$ is the calculated INS down acceleration component and ${\nu}_{_{\textbf{ZAD}}}$ is zero mean white Gaussian measurement noise. Under the navigation filter setup, the accelerometer residual satisfies $\delta \boldsymbol{f}_{ib}^b \approx \mathbf{b}_{a}$, such that the measurement matrix is defined by:
\begin{equation}
\textbf{H}_{_{\textbf{ZAD}}} = \begin{bmatrix}
\textbf{0}_{1 \times 3} & \textbf{0}_{1 \times 3} & \textbf{0}_{1 \times 3} & \textbf{e}^{\operatorname{T}}_3 & \textbf{0}_{1 \times 3}
\end{bmatrix}.
\end{equation} 
\textbf{In the Literature}: As an extension to the constant height constraint (\ref{const_height}), zero vertical body velocity must lead to zero down acceleration in the body frame. Klein used this vehicle constraint to aid the INS navigation in urban environments, where the surface is typically level \cite{klein2010pseudo, klein2011mitigating, klein2011vehicle}. 
\subsection{Zero Acceleration in Navigation Frame (ZAN)} \label{zero_acc_navigation}
\noindent
\textbf{Assumption}: Stationary conditions. \\
\textbf{Applications}: Land vehicles and stationary fine alignment. \\
\textbf{Principle}: While stationary, the acceleration is zero and thus the accelerometer measurements expressed in the navigation frame are given by: 
\begin{equation}
\boldsymbol{f}^n = \textbf{T}_b^n \boldsymbol{f}_{ib}^b = \textbf{g}^n .
\end{equation}
\textbf{Measurement Model}: The measurement residual is given by: 
\begin{align}\label{eq:ZAN1}
\delta \mathbf{z}_{_{\textbf{ZAN}}} &= \boldsymbol{f}^n_{_{\text{INS}}} - \boldsymbol{f}^n = \textbf{H}_{_{\textbf{ZAN}}} \delta \mathbf{x} + \boldsymbol{\nu}_{_{\textbf{ZAN}}}
\end{align} 
where $\boldsymbol{f}^n_{_{\text{INS}}}$ is the specific force measured in the navigation frame,  $\boldsymbol{\nu}_{_{\textbf{ZAN}}}$ is zero mean white Gaussian measurement noise, and $\textbf{H}_{_{\textbf{ZAN}}}$ is the measurement matrix. It is obtained by perturbing the measurement residual (\ref{eq:ZAN1}) 
\begin{align}
\delta \boldsymbol{f}^n &= \Tilde{\boldsymbol{f}}^n - \boldsymbol{f}^n = 
(\Tilde{ \textbf{T}}_b^n \Tilde{\boldsymbol{f}}^b ) - \boldsymbol{f}^n \notag \\
&= \Big( \textbf{T}_n^b  (\textbf{I} - [ \boldsymbol{\epsilon}^n \times ] )   \boldsymbol{f}^b + \textbf{T}_b^n \delta \boldsymbol{f}^b \Big) - \boldsymbol{f}^n \notag \\
&= [ \boldsymbol{f}^n \times ] \boldsymbol{\epsilon}^n + \textbf{T}_n^b \textbf{b}_a \end{align}
such that the measurement matrix is:
\begin{equation}
\textbf{H}_{_{\textbf{ZAN}}} = \begin{bmatrix}
\textbf{0}_{3 \times 3} & \textbf{0}_{3 \times 3} & [\boldsymbol{f}^n \times ] & \textbf{T}_b^n & \textbf{0}_{3 \times 3}
\end{bmatrix}.
\end{equation} 
\textbf{In the Literature}: Farrell used ZAN for a vehicular implementation where small angles were replaced with a known local gravity vector in the n-frame \cite{farrell2008aided}, Klein used it as an external measurement in the navigation filter to improve attitude accuracy during stationary fine alignment \cite{klein2021ins}.
\subsection{Constant LLH Position (CP)} \label{const_position} 
\noindent
\textbf{Assumption}: Slow motion in urban environments for short time periods. \\
\textbf{Applications}: Land vehicles and calibration. \\
\textbf{Principle}: In some cases, the change in latitude and longitude can be assumed as zero. That is, for short time periods and relatively low velocities, the displacement is rather small and therefore assumed to be zero. \\
\textbf{Measurement Model}: The measurement residual is given by:
\begin{equation}
\delta \mathbf{z}_{_{\textbf{CP}}} = \textbf{p}^n_{_{\text{INS}}} - \textbf{p}^n_c = \textbf{H}_{_{\textbf{CP}}} \delta \mathbf{x} + \boldsymbol{\nu}_{_{\textbf{CP}}}
\end{equation}
where $\textbf{p}^n_{\text{C}}$ is the last known position, $\textbf{p}^n_{\text{INS}}$, is the calculated INS position, and $\boldsymbol{\nu}_{_{\textbf{CP}}}$ is zero mean white Gaussian measurement noise. The measurement matrix is defined by:
\begin{equation}
\textbf{H}_{_{\textbf{CP}}} = \begin{bmatrix}
\textbf{I}_{3 \times 3} & \textbf{0}_{3 \times 3} & \textbf{0}_{3 \times 3} & \textbf{0}_{3 \times 3} & \textbf{0}_{3 \times 3}
\end{bmatrix}.
\end{equation} 
\textbf{In the Literature}: 
Klein et al. (2010) used it for land vehicles, demonstrating significant improvement of the estimation error, compared to other aiding types \cite{klein2010pseudo}. Li (2012) \cite{li2012situ} proposed a quick hand calibration method where pseudo-position and pseudo-velocity observations were used as a substitute for missing GPS information. Ilyas (2016) \cite{ilyas2016drift} proposed a position and attitude (P-A) locking mechanism for PDR applications. Once long standstill conditions are met by the motion detector algorithm, both parameters get the last measurements obtained before the pedestrian motion ceases.
\subsection{Zero Yaw Rate (ZYR)} \label{zero_yaw_rate} \noindent
\textbf{Assumption}: Motion at zero yaw rate. \\
\textbf{Applications}: Land vehicles, mobile robots, and pedestrian navigation. \\
\textbf{Principle}: Given no road bank or slope, motion in straight lines results in a zero rate of change of yaw angle. \\
\textbf{Measurement Model}: The measurement residual is given by:
\begin{equation}
\delta \mathbf{z}_{_{\textbf{ZYR}}} = \omega_{ib,z_{\text{ INS}}}^b - 0 = \textbf{H}_{_{\textbf{ZYR}}} \delta \mathbf{x} + {\nu}_{_{\textbf{ZYR}}}
\end{equation}
where $\omega_{ib,z_{\text{ INS}}}^b$ is the calculated INS yaw rate and  ${\nu}_{_{\textbf{ZYR}}}$ is zero mean white Gaussian measurement noise. The measurement matrix is defined by:
\begin{equation}
\textbf{H}_{_{\textbf{ZYR}}} = \begin{bmatrix}
\textbf{0}_{1 \times 3} & \textbf{0}_{1 \times 3} & \textbf{0}_{1 \times 3} & \textbf{0}_{1 \times 3} & \textbf{e}^{\operatorname{T}}_3
\end{bmatrix}.
\end{equation}
\textbf{In the Literature}: Crasta \cite{crasta2015observability} examined ZYR from an observability analysis standpoint of a 3D autonomous underwater vehicle (AUV) in the presence of ocean currents. Huttner used ZYR \cite{huttner2018offset} for a real-time estimate of longitudinal and lateral acceleration offsets. Niu \cite{niu2010using} used ZYR in low-speed dynamics as land vehicles are subject to steering constraints. 

\subsection{Constant Heading Angle (CHA)} \label{const_heading_angle}
\noindent
\textbf{Assumption}: Motion at constant heading angle. \\
\textbf{Applications}: Land vehicles, mobile robots, and pedestrian navigation. \\
\textbf{Principle}: In urban environments with grid-based street layout, motion in short time intervals can be said to be rectilinear, i.e., the heading angle does not change, $\psi = \psi_c$. \\
\textbf{Measurement Model}: Assuming small heading angles, the measurement residual is given by:
\begin{equation}
\delta {z}_{_{\textbf{CHA}}} = \psi_{_{\text{INS}}} - \psi_c = \textbf{H}_{_{\textbf{CHA}}} \delta \mathbf{x} + {\nu}_{_{\textbf{CHA}}}
\end{equation}
where $\psi_{_{\text{INS}}}$ is the calculated INS heading angle and ${\nu}_{_{\textbf{CHA}}}$ is zero mean white Gaussian measurement noise. The measurement matrix is defined by:
\begin{equation}
\textbf{H}_{_{\textbf{CHA}}} = \begin{bmatrix}
\textbf{0}_{1 \times 3} & \textbf{0}_{1 \times 3} & \textbf{e}^{\operatorname{T}}_3 & \textbf{0}_{1 \times 3} & \textbf{0}_{1 \times 3}
\end{bmatrix}.
\end{equation}
\textbf{In the Literature}: The constant heading angle, also known as the zero integrated heading rate (ZIHR), was mentioned by Shin \cite{Shin2001, shin-Thesis, shin-PhD} as a means of restricting the heading drift during vehicle stops, and Liu \cite{liu2012performance} determined CHA using the last-stored heading angle from the vehicle stop. 
In pedestrian navigation applications, Abdulrahim \cite{abdulrahim2012using} noted several rules of thumb to constrain heading angles: $i)$ Building shapes are mostly rectangular, forcing pedestrians to walk in parallel to only one of the four sides at a time. $ii)$ Non-walking situations should be identified to restrict heading drift during  conditions. $iii)$ Turnings occur to a small extent, exhibit sharp signals, and for small change of direction. Borenstein \cite{borenstein2009heuristic-1, borenstein2009heuristic-2, borenstein2010heuristics} and later Jiménez \cite{jimenez2011improved} introduced heuristic drift reduction (HDR) to update CHA as most indoor walking is performed along straight pathways. Additional works incorporated biomechanical models \cite{alvarez2012pedestrian, zhang2020improved} and others employed heuristics/fuzzy logic methods to determine CHA by averaging the last n-footsteps in a stance phase \cite{ilyas2016drift, bousdar2017loose, ahmed2018wearable, deng2018improved}.
\subsection{Virtual Lever-Arm (VLA)} \label{lever_arm}
\noindent
\textbf{Assumption}: Motion on a planar surface. \\
\textbf{Applications}: Any platform with an INS and a GNSS receiver. \\
\textbf{Principle}: The basic idea behind VLA measurement is that the lever arm (LA) is known to some accuracy level; however, this knowledge is not used directly in the navigation filter, but mostly utilized to model the stochastic process describing the lever arm error characteristics. By introducing VLA we utilize this knowledge directly to improve the navigation performance. To that end, first the lever-arm error states are augmented with \eqref{eq_iInsErrorState} resulting in
\begin{equation}\label{eq:iInsErrorState18}
\delta \mathbf{x}_{_{\textbf{VLA}}} = \left[ \begin{array}{cc} 
\delta \mathbf{x} & \delta \boldsymbol{l}^b \end{array} \right]^{\operatorname{T}}\in\mathbb{R}^{18}
\end{equation}
where $\delta \boldsymbol{l}^b \in \mathbb{R}^3$ is the lever arm error state vector. \\
\textbf{Measurement Model}: The measurement residual is given by: 
\begin{equation} \label{eq:VLA}
\delta \mathbf{z}_{_{\textbf{VLA}}} = \textbf{v}_{l,_{\text{INS}}}^n - \textbf{v}^n = \textbf{H}_{_{\textbf{VLA}}} \delta \mathbf{x}_{_{\textbf{VLA}}} + \boldsymbol{\nu}_{_{\textbf{VLA}}}
\end{equation}
where $\textbf{v}_{l,_{\text{INS}}}^n$ is the calculated INS velocity measurement containing a lever arm and $\boldsymbol{\nu}_{_{\textbf{VLA}}}$ is zero mean white Gaussian measurement noise. 
The LA-aided velocity term (\ref{eq:VLA}) is a function of several parameters that undergo small error perturbation analysis, resulting in the following measurement residual:
\begin{align*} 
\delta \textbf{v}_l^n = \Tilde{\textbf{v}}_l^n - \textbf{v}_l^n = \delta \textbf{v}^n + \Big( (\textbf{I} + [\boldsymbol{\epsilon}^n \times]) \textbf{T}_b^n [ \boldsymbol{\omega}_{ib}^b \times] \, \boldsymbol{l}^b \hspace{1cm} \\
- \textbf{T}_b^n [\boldsymbol{l}^b \times] ( \boldsymbol{\omega}_{ib}^b + \delta \boldsymbol{\omega}_{ib}^b ) \,  + \textbf{T}_b^n [\boldsymbol{\omega}_{ib}^b \times] \, (\boldsymbol{l}^b + \delta \boldsymbol{l}^b) \Big) - \textbf{v}_l^n  \\
= \delta \textbf{v}^n - [\textbf{T}_b^n (\boldsymbol{\omega}_{ib}^b \times) \boldsymbol{l}^b \times] \boldsymbol{\epsilon}^n - \textbf{T}_b^n [\boldsymbol{l}^b \times] \textbf{b}_g + \textbf{T}_b^n [\boldsymbol{\omega}_{ib}^b \times] \, \delta \boldsymbol{l}^b
\end{align*}
denote $[\boldsymbol{\omega} \times]$=$\Omega$ and $[\boldsymbol{l} \times]$=$\boldsymbol{L}$, such that the measurement matrix is given by:
\begin{equation}
\textbf{H}^v_{_{\textbf{VLA}}} = \begin{bmatrix}
\textbf{0}_{3 \times 3} \hspace{2mm} \textbf{I}_{3 \times 3} \, - ( \textbf{T}_b^n \boldsymbol{\Omega}_{ib}^b \boldsymbol{l}^b \times) \hspace{2mm} \textbf{0}_{3 \times 3}  - \textbf{T}_b^n \boldsymbol{L}^b \hspace{2mm} \textbf{T}_b^n \boldsymbol{\Omega}_{ib}^b
\end{bmatrix}.
\end{equation}
\textbf{In the Literature}: Borko et al. (2018) \cite{borko2017observability, borko2018gnss} formulated the VLA measurement as a dynamical state variable, thus demonstrating a significant improvement in both accuracy and error-states convergence time. Monaco (2018) \cite{monaco2018doppler} showed that when the DVL unit is mounted far from the AUV's center of buoyancy, the lever-arm offset improves the vehicle's states observability. Zhang (2020) \cite{zhang2020required, zhang2021mounting} investigated the effect of lever-arm error on the contribution of the information-aided measurements (non-holonomic constraints) for land vehicle navigation. Recently, Hwang (2021) \cite{hwang2021identification} proposed a modified single track model to decouple the lever-arm effect from the vehicle dynamic states using an UKF.
\\
\begin{figure}[!t]
\begin{center}
\includegraphics[width=0.4\textwidth]{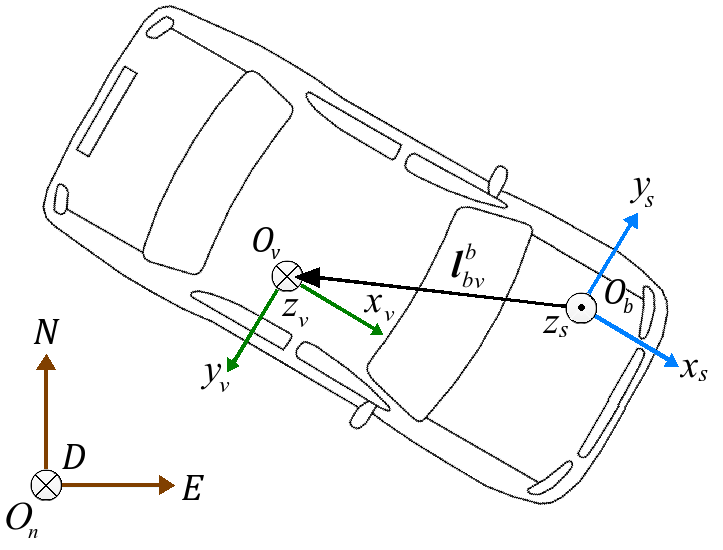}
\caption{Schematics of the lever arm between the IMU and vehicle frames}
\label{f:lever-arm}
\end{center}
\end{figure}
\subsection{Relative Update (RUP)} \label{m_imus} \noindent
\textbf{Assumption}: Lever-arm vector between multiple IMUs is known a priori.\\
\textbf{Applications}: Any type of platform with multiple IMUs. \\
\textbf{Principle}: When multiple IMUs are fused together, each block filter provides its own position, velocity, and attitude estimates:
\begin{align} \label{eq:multiple}
\begin{bmatrix} 
\delta\dot{\mathbf{x}}_{\text{k+1}}^1 \\ \vdots \\ \delta\dot{\mathbf{x}}_{\text{k+1}}^n
\end{bmatrix} = \begin{bmatrix} 
\mathbf{\Phi}_{\text{k,k+1}}^1 & \hdots & 0 \\
\vdots & \ddots & \vdots \\
0 & \hdots & \mathbf{\Phi}_{\text{k,k+1}}^n \end{bmatrix} \begin{bmatrix} 
\delta\mathbf{x}_{\text{k}}^1 \\ \vdots \\ \delta\mathbf{x}_{\text{k}}^n
\end{bmatrix} + 
\begin{bmatrix}
\delta\mathbf{w}_{\text{k}}^1 \\ \vdots \\ \delta\mathbf{w}_{\text{k}}^n
\end{bmatrix}
\end{align}
Since IMUs are rigidly mounted, distance vectors between them remain constant in the body frame, allowing relative information to be used as updates.\\
\textbf{Measurement Model}: The measurement residual provides the following relative position update between every two IMUs:
\begin{equation}
\delta \mathbf{z}_{_{\textbf{RUP}}}^{\text{P}} = \textbf{p}^{b,1}_{_{\text{INS}}} - \textbf{p}^{b,2}_{_{\text{INS}}} - \textbf{p}^{2,1}_{_{\text{INS}}} = \textbf{H}_{_{\textbf{RUP}}}^{\text{P}} \delta \mathbf{x} + \boldsymbol{\nu}_{_{\textbf{RUP}}}^{\text{P}}
\end{equation}
where $\textbf{p}^{b,1}_{_{\text{INS}}}$ is the position estimates of the first block filter and $\textbf{p}^{2,1}_{_{\text{INS}}}$ is the known relative distance vector between both IMUs. The corresponding process noise of each block filter is given by zero mean white Gaussian noise $\boldsymbol{\nu}_{_{\textbf{RUP}}}^{\text{P}}$ and the measurement matrix is defined by:
\begin{equation}
\textbf{H}_{_{\textbf{RUP}}}^{\text{P}} = \begin{bmatrix}
\textbf{I}_{3 \times 3} & \textbf{0}_{3 \times 12} \hspace{2mm} \big| \ -\textbf{I}_{3 \times 3} & \textbf{0}_{3 \times 12}
\end{bmatrix}.
\end{equation}
By using the rotation rate vector as measured by the first IMU, a relative velocity vector $\dot{\textbf{p}}^{2,1}=\boldsymbol{\omega}_1 \times \textbf{p}^{2, 1}_{_{\text{INS}}}$ enables generating a relative velocity update, given by the measurement matrix:
\begin{align}
\textbf{H}_{_{\textbf{RUP}}}^{\text{V}} = 
\begin{bmatrix}
\textbf{0}_{3 \times 3} & \textbf{I}_{3 \times 3} & \textbf{0}_{3 \times 9} \hspace{2mm} \big| \ \textbf{0}_{3 \times 3} -\textbf{I}_{3 \times 3} & \textbf{0}_{3 \times 9} 
\end{bmatrix}.
\end{align}
And since relative orientation between fixed IMUs also remains constant during motion, the relative attitude update is given by the following measurement matrix:
\begin{align}
\textbf{H}_{_{\textbf{RUP}}}^{\text{A}} = \begin{bmatrix}
\textbf{0}_{3 \times 6} & \textbf{I}_{3 \times 3} & \textbf{0}_{3 \times 6} \hspace{2mm} \big| \ \textbf{0}_{3 \times 6} -\textbf{I}_{3 \times 3} & \textbf{0}_{3 \times 6}
\end{bmatrix}.
\end{align}
\textbf{In the Literature}: Bancroft (2008) \cite{bancroft2008twin, bancroft2010multiple, bancroft2011data} introduced several foot-mounted architectures for pedestrian INS, where multiple MEMS IMUs were integrated into a common virtual frame (VIMU), using knowledge of the human gait cycle for relative updates. Waegli (2008) \cite{waegli2008redundant} investigated how measurements obtained from redundant sensor constellations at different geometries affect the estimates accuracy. Guerrier (2009) \cite{guerrier2009improving} showed that an IMU in a triad configuration exhibits invariance to the geometrical orientation between them. Next, vehicular implementation was introduced, where relative distance vectors were obtained by differencing the lever arms between sensor assembly and the GNSS antenna \cite{bancroft2009multiple}. Recently, Hartzer (2020) \cite{hartzer2020decentralized} introduced a multi-IMU distributed over several autonomous vehicles, forming a collaborative localization environment using relative position and attitude updates from their GNSS antennae.

\section{Indirect Information Aiding} \label{sec_4} \noindent
In Section~\ref{sec_3}, platform constraints and information are used directly for filter updates. In some situations, more information can be squeezed from external sensors, hence they are addressed as indirect information aiding. Table~\ref{tab:IB} summarizes the different indirect information aiding methods reviewed in this section, broadly categorized into two families of modalities: 
\begin{table}[!h] 
\centering
\caption{Indirect information aiding types} \label{tab:IB}
\begin{tabular}{ >{\centering}p{7mm} >{\centering}p{10mm} p{18mm} >{\centering}p{20mm} >{\centering}p{12mm} }
 Section & Abbrv. & \hspace{0mm} Aiding Type & Information Type & Modality \\ [2mm] \hline \\
\ref{PbO} & PBO & Position-based & \multirow{2}{*}{Orientation} & \multirow{2}{*}{\vtop{\hbox{\strut Hetero-}\hbox{\strut \ geneous}}} \\ [1.50mm] 
\ref{vbh} & VBO & Velocity-based & & \\ [1.50mm] \hline \\[.2mm] 
\ref{PG-GNSS} & P-GNSS & Pseudo-GNSS & \multirow{2}{*}{Sensor} & \multirow{2}{*}{\vtop{\hbox{\strut Homo-}\hbox{\strut \ geneous}}} \\ [1.50mm]
\ref{BV-DVL} & P-DVL & Pseudo-DVL & & \\ [1.50mm]
\hline \end{tabular} 
\end{table}\\
i) Heterogeneous: quantities from one state variable (provided from external sensors) provide new information about the other, e.g., position measurements are used to infer the platform attitude. ii) Homogeneous: same quantities obtained by the external sensors are later approximated by model-based or learning-based approaches, e.g., pseudo-sensor information enables the use of the external sensor.\\
\begin{figure}[h]
\begin{center}
\includegraphics[width=0.485 \textwidth]{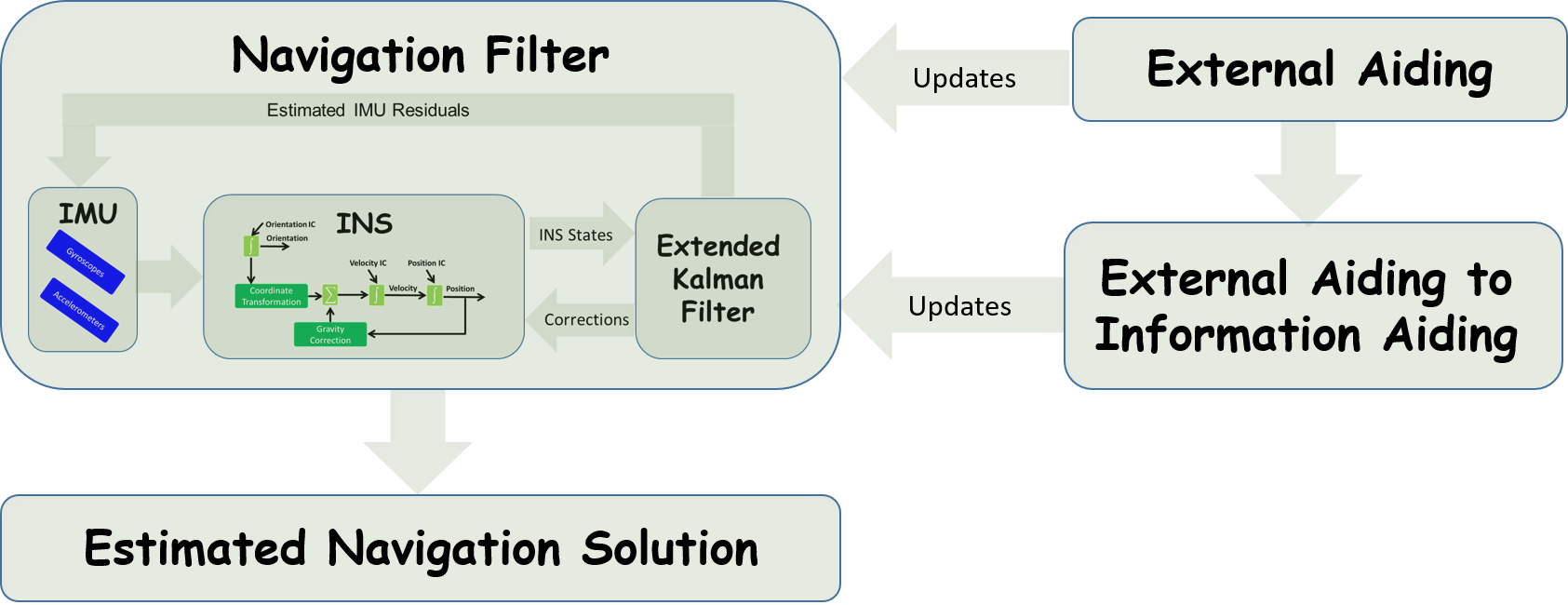}
\caption{Indirect information aiding block diagram.}
\label{f:inaIA}
\end{center}
\end{figure}\\
Note that direct information aiding does not require external sensors, yet heterogeneous indirect information aiding does require external sensor, as shown in Figure~\ref{f:inaIA}.
\subsection{Position-Based Orientation (PBO)} \label{PbO} \noindent
\textbf{Applications}: Vehicles in any operational environment.\\
\textbf{Use case}: Low accelerations, constant velocity, and especially stationary conditions often lead to weak observability of some error states such as misalignment angles or instrumental biases. Without direct angle measurements, the orientation error may evolve and expedite the navigation error due to gravity misprojection.\\
\textbf{Principle}: The kinematic trajectory of the vehicle over time provides an abundance of geometric and temporal constraints. Exploiting these relations provides not only orientation-related information, but also unobservable error states.\\
\textbf{In the literature}: Cho (2010) \cite{cho2010performance} and Maeder (2011) \cite{maeder2011attitude} devised a novel technique that provides additional attitude information to GPS/INS integrated systems. Wu (2012) \cite{wu2012improving} used a cascaded Kalman filter to filter the GPS course angle and aid the unobservable yaw angle. Using an acceleration-sensitive switching mechanism, the divergence of the yaw angle was reduced during horizontal constant speed. Juang (2014) \cite{juang2014sensor} used a difference-based approach where previous estimates of the position in plane were used to approximate the speed and the heading angle.\\
Wierzbicki (2015) \cite{wierzbicki2015estimation} showed how GPS signals can be used as inputs for the Helmert transformation, thus enabling the determination of the rotation angles using relative changes in the geocentric reference frame. Li (2016) \cite{li2016comparison} examined the performance of GPS-aided attitude estimates where a direct navigation approach was compared with an indirect error state, showing a clear advantage to the latter. Huang (2018) \cite{huang2018new, huang2020high} introduced a GPS-aided coarse alignment method where the attitude matrix between initial and current body frames is jointly inferred based on the derived linear state-space model using KF. Xu \cite{xu2019motion} proposed an efficient misalignment estimation for the SINS/GPS integrated system using position loci. Klein (2020) \cite{klein2020squeezing} introduced an approach where previously measured position updates were simultaneously utilized in the navigation filter to determine the vehicle heading. 

\subsection{Velocity-based Orientation (VBO)}  \label{vbh} \noindent
\textbf{Applications}: Land vehicles in motion and aerial vehicles under the assumption of coordinated flight. \\
\textbf{Use case}: Some navigational scenarios may suffer from poor performance due to unmodelled dynamics that interfere with the motion model. For example, winds or sea currents may cause unwanted lateral velocities, resulting in a side-slip angle between the vehicle pointing direction, i.e., yaw angle, and the actual travel direction, i.e., course angle.\\
\textbf{Principle}: The rotation of the velocity coordinate frame with respect to the navigation frame (NED) can be used for spatial constraints, providing additional attitude information. \\
\textbf{In the literature}: Pseudo-attitude dates back to the 90's where Cohen \cite{cohen1992aircraft} and Kornfeld \cite{kornfeld1998single} showed how decomposition of the velocity and acceleration vectors can be manipulated to improve the angle estimates. Shin (2001) \cite{Shin2001} used a velocity matching alignment, where GPS measurements and aircraft dynamics were synthesized to allow angle estimates from navigation frame velocities.
Salycheva (2004) \cite{salycheva2004kinematic} proposed a threshold mechanism that is activated once maneuvers are detected. Using both GPS velocity and heading information to allow azimuth error estimation, the misalignment error was shown to become a directly measurable component. Bevly (2006) \cite{bevly2006integrating} proposed a unique method to determine attitude and sideslip angle using velocity measurements.
Jie (2008) \cite{jie2008design} proposed decomposing both velocity and acceleration vectors into tangential and normal components, such that new reference vectors are formed in the horizontal and vertical planes, enabling the extraction of Euler angles.\\
Lai (2010) \cite{lai2010application} proposed estimating a pseudo-roll angle from the aircraft velocity vector axis to compensate for influences of the angle of attack in the longitudinal plane. Wu and Pan (2013) \cite{wu2011optimization, wu2013velocity-1, wu2013velocity-2} proposed a closed-form solution to extract the orientation matrix in scenarios where airborne dynamics are likely to interfere using position/velocity inputs. Wang \cite{wang2017attitude} fused the pseudo-attitude approach using low-cost MEMS IMU, demonstrating a significant improvement in accuracy compared to unaided dead reckoning.
\subsection{Pseudo-GNSS (P-GNSS)} \label{PG-GNSS} \noindent
\textbf{Applications}: Vehicles in any operational environment.\\
\textbf{Use case}: Unstable GNSS reception can be caused by interference, spoofing, or signal blockage due to environmental constraints. Unlike the tightly coupled (TC) INS/GNSS integration strategy, the loosely coupled approach cannot handle scenarios with less than four available satellites \cite{grewal2007global}.\\ 
\textbf{Principle}: Position increments are highly correlated with vehicle dynamics, due to their kinematic coupling. To exploit this, both analytical models and learning-based approaches were proposed, enabling regression of pseudo-GNSS increments to substitute for the missing signals.\\
\textbf{In the literature}: 
Conley (2006) \cite{conley2006performance} introduced a fictitious-GPS approach where pseudo-satellite signals were obtained from minimum Dilution of precision (DOP), and each initial position and velocity is determined using almanac data. El (2006) \cite{el2006utilization} and Noureldin (2011) \cite{noureldin2011gps} were among the first to train neural networks to map position and velocity error with GPS corrections.
Klein (2011) \cite{Klein2011, klein2012odometer} proposed a modified loosely coupled (MLC) approach where virtual satellites provide complementary pseudorange measurements, derived mathematically from the INS position and velocity estimates.\\
Tan (2015) \cite{tan2015ga} showed that machine learning models combined with loosely coupled integration schemes are capable of predicting positions once trained on GPS increments that are highly correlated with specific forces and angular rate measurements. Yao, Zhang, and Fang (2017) \cite{yao2017rls, yao2017hybrid, zhang2019fusion, fang2020lstm} introduced different state machines where learning-based models were trained on pairs of error vectors and their corresponding position increments during normal operation. Once GNSS signals ceased to be received, a switching mechanism activates inference mode, where models regress pseudo-GNSS signals.
\subsection{Pseudo-DVL (P-DVL)} \label{BV-DVL} \noindent
\textbf{Applications}: Marine environment.\\
\textbf{Use case}: Doppler velocity log (DVL) is a common aiding sensor at subsea environment where satellite reception is unavailable and no pre-installed infrastructure is required. Given sensitivity to the acoustic environment or proximity to the seafloor, some DVL beams are not reflected back and the DVL fails to maintain bottom lock, being unable to estimate the AUV velocity.\\
\textbf{Principle:} Additional information or sensor measurements are used to form and estimate the missing DVL beams. Those beams, together with the measured available beams, are combined to estimate the velocity vector of the vehicle.\\
\textbf{In the literature}: The works are divided into two families of constraints from which the information is derived. 
\begin{itemize}
    \item \textbf{Geometric constraints}: exploit spatial relations that exist between different state variables.
    \item \textbf{Temporal constraints}: exploit the vehicle's kinematic trajectory over time for the purpose of obtaining complementary information.
\end{itemize}
\subsubsection{\textbf{Geometric constraints}}
Zhu (2017) \cite{zhu2017novel} employed a partial least squares regression combined with support vector regression (SVR) to form a hybrid predictor, mapping the missing DVL measurements to current and past INS velocity estimates. Eliav (2018) \cite{eliav2018ins} also used an ELC integration approach that combines partial DVL measurements with the previous navigation solution to form calculated velocity estimates to aid the INS. That is, past DVL velocity measurement and inertial sensor readings are used together with partial beam measurements to form an estimation of the vehicle velocity vector.\\
Liu (2018) \cite{liu2018ins} implemented tightly coupled navigation approach, where depth updates from a pressure sensor were used to complement limited DVL measurements via geometrical constraints. Similarly, Wang (2019) \cite{wang2019novel} used tightly integrated approach combined with a pressure sensor, but used raw beam measurements without transforming them into a 3D velocity vector. Ansari \cite{ansari2019pseudo} showed how missing DVL data can be predicted using evolutionary fuzzy algorithm, without any auxiliary sensor information. Yoo \cite{yoo2019design} implemented a TC integration scheme which is capable of using the partial DVL measurements by regenerating the measurements of the failed transducers.\\
Li (2020) \cite{li2020novel, li2021underwater} presented a novel neural network-based SINS/DVL which is able to provide reliable velocity forecasts during long DVL malfunctions. Wang (2021) \cite{wang2021virtual} implemented the Dempster-Shafer theory augmented by least squares SVM to synthesize virtual DVL signals. Recently, Yao (2022) \cite{yao2022virtual} introduced a ZUPT-aided virtual beam solution, where geometrical relationship were employed between the DVL beams and the zero velocity vector. Yona, and later Cohen (2022) \cite{yona2021compensating, cohen2022set, cohen2022libeamsnet}, used dedicated neural convolutional networks to regress the missing beam velocity, given scenarios of sensor malfunctions, partial beam measurements, or outliers. 

\subsubsection{\textbf{Temporal constraints}} Stanway (2012) \cite{stanway2012contributions} used online bin-averaging to estimate vehicle velocities in surface denied regions below 200 m, based on the weighted average difference with respect to past observations. Ivanov (2014) \cite{ivanov2014resilient} used polyhedrons to model the time-varying trajectory as a convex set of vertices in 3D space. A resilient sensor fusion algorithm was shown to use the platform history not only for navigation purposes but also for protection against sensor attacks or failures. Costanzi (2017) \cite{costanzi2017evaluation, tesei2018real, costanzi2018estimation} introduced different cooperative strategies to handle the DVL bottom lock range in deep waters, using an online-updated database of past measurements and estimates by multiple EKF carriers.\\
The set of papers \cite{klein2020continuous, klein2021insdrift, klein2022estimating} addressed scenarios of partial and complete DVL beam outage, using analytically derived algorithm that estimates the velocity vector and its corresponding variance. By replacing simple averaging with higher order polynomial approximation, derivatives of the past estimated velocities were accounted for, enabling the DVL acceleration and the jerk term to be estimated as well. 

\section{Model-Aided INS} \label{sec_5} 
\begin{figure}[!h]
\begin{center}
\includegraphics[width=0.5 \textwidth]{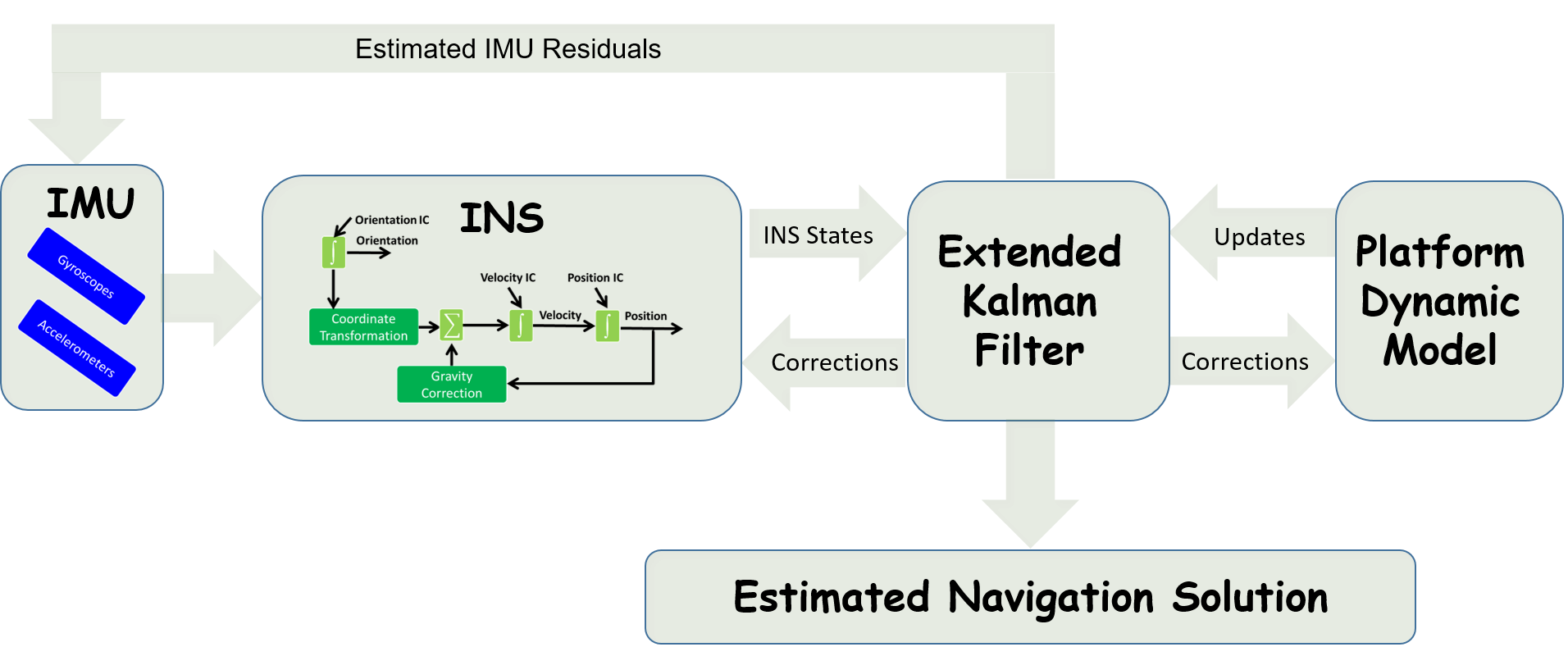}
\caption{Model aided information block diagram.}
\label{f:modela}
\end{center}
\end{figure}

\noindent 
A vehicle dynamic model (VDM), mostly at six degree of freedom (DoF), is a simulation of a vehicle capable of generating specific force and angular velocity vectors, induced by the forces and moments acting on the vehicle. Table~\ref{tab:IC} summarizes the different works found in the literature categorized according to the specific operational environment.
\begin{table}[!h] 
\centering
\caption{Model-based aiding types} \label{tab:IC}
\begin{tabular}{ >{\centering}p{0.7cm} >{\centering}p{1.2cm} >{\centering}p{3.5cm} }
Section  & Abbrv. & \hspace{2mm} Operational Environment \\ [2mm] \hline \\
\ref{aerial} & AV & Aerial \\ [1.50mm] \hline \\[.2mm]
\ref{marine} & MV & Marine \\ [1.50mm] \hline \\[.2mm]
\ref{Underwater} & UV & Underwater \\ [1.50mm] \hline \\[.2mm]
\ref{land_vehicles} & LV & Land \\ [1.50mm]
\hline \end{tabular} 
\end{table}\\
While their errors are attributed to errors in the platform model such as inaccurate drag coefficients of wind model, the measured inertial readings also contain errors associated with the inertial hardware and environmental conditions. As the error comes from two different source (platform modeling and inertial hardware), the simulated motion model runs in parallel to the navigation filter, feeding each other reciprocally, thus avoiding the standalone INS operation.\\
The model-aided INS can be applied without external sensor, but requires sufficient computational capabilities to execute the 6-DoF motion model running online. Beside being an additional information source, this concept also offers powerful physical predictions, e.g., aerodynamic or hydrodynamic effects that are otherwise imperceptible by the sensor array; that is, the inertial readings help to correct the simulation parameters. Figure~\ref{f:modela} illustrates the connection configuration of the VDM module with the INS solution.
%

\subsection{Aerial Vehicles (AV)} \label{aerial} \noindent 
Koifman and Bar-Itzhack (1999) \cite{Koifman1999} introduced an aircraft dynamic model (ADM) as an aid for a low-grade strapdown INS. Given constant wind velocity, the dynamics-aided solution was shown to be significantly more accurate than that of an unaided INS. Julier (2003) \cite{julier2003role} examined the role played by complementary vehicle models and their impact on the performance of sensor-based navigation systems. Bryson (2004) \cite{bryson2004vehicle} implemented the ADM approach to aid pose estimates for unmanned aerial vehicles (UAV).\\
Cork (2007) \cite{cork2007sensor} presented a nonlinear model of the aircraft dynamics using an unscented Kalman filter (UKF) alongside sensor fault detection. Vissiere (2008) \cite{vissiere2008experimental} used an extensive environmental model that accounts for ground effects, actuator-induced lags, rotor-induced body drag, and flexible responses of a small-scale helicopter motion model. Dadkhah \cite{dadkhah2008model} used equations of motion as an aiding source for an AHRS in GPS-denied environments, using a miniature helicopter model.
Crocoll (2013) \cite{crocoll2013unified, crocoll2014model, crocoll2014quadrotor} combined both the kinematic model with the VDM into an optimal unified model, thus managing to outperform existing methods.
Cork (2014) \cite{cork2014aircraft} incorporated the ADM directly into the navigation filter of a fixed-wing UAV, using the emulated forces and moments acting on it as information sources. Butt (2015) \cite{butt2015adaptive} presented a hypersonic flight vehicle modeling (FVM), validated against an actual NASA flight test.\\
Khaghani (2016) \cite{khaghani2016evaluation, khaghani2018vdm} used a detailed wind model alongside the AUV's process model, exhibiting dramatic improvements of the navigation filter. Nobahari (2017) \cite{nobahari2017application} introduced model-aided inertial navigation (MAIN), based on a 6-DoF flight simulation that acts as an INS aiding means during the landing phase or ground proximity. Mwenegoha (2019) \cite{mwenegoha2019model} implemented a model-based integration approach for a fixed-wing UAV, using VDM aided by low-cost MEMS inertial sensors and GNSS measurements with moment of inertia calibration. Laupre (2020) \cite{laupre2020self} showed how the VDM is strongly dependent on the accurate determination of the aerodynamic coefficients. Youn \cite{youn2020model, youn2020aerodynamic, youn2021model} calculated aerodynamic coefficients in real-time, to synthesize control signals, improving attitude estimates compared to conventional approaches that do not account for wind models. Similarly, Ko, and later Xu (2019) \cite{ko2019three, xu2021vehicle}, proposed a practical VDM reconstruction method to aid quadrotors during GPS outages.
\subsection{Marine Vehicles (MV)} \label{marine} \noindent 
Shi (2009) \cite{shi20092} proposed a nonlinear model frame of a maneuvering ship based on an analysis of the hydrodynamics and preprocessed by a fixed interval KF smoothing algorithm. Perera (2010) \cite{perera2010ocean, perera2017navigation} presented a maneuvering ocean vessel model based on a curvilinear motion model, based on a linear position model for accurate trajectory estimation. Fossen \cite{fossen2009kalman} introduced a marine motion control unit that uses KF to reduce oscillatory wave-induced motion from velocity and position measurements and improve the heading angle. Soda (2011) \cite{soda2011simulation} accounted for several environmental factors that interact with the vessel's dynamics, e.g., ocean currents, waves, and tides, and used them as inputs to the ship navigation states. \\
Vasconcelos \cite{vasconcelos2011ins} used gravitational-aided observations, by deriving a sensor integration scheme that accounts the dynamics of the autonomous surface craft (ASC) to properly trace measurement disturbances. Araki (2012) \cite{araki2012estimating} incorporated computational fluid dynamics (CFD) analysis in the motion model, to better predict the ship maneuverability. 
Chen (2013) \cite{chen2013numerical, chen2015effect} conducted numerical simulations to examine how waves, ocean currents, and winds affect navigation performance in terms of drift, based on the Princeton Ocean Model (POM). Sabet (2017) \cite{sabet2017identification} extended the 6-DoF motion model by parameterizing viscous damping, body lift, and control inputs, using cubature and an unscented KF. Lu (2018) \cite{lu2018ocean} presented a dynamically derived model for ocean vehicles, which incorporates nonholonomic constraints (NHC) for a maritime environment in simulation engine. Taimuri (2020) \cite{taimuri20206} introduced a mathematical model of propulsion ships, where the heave, roll, and pitch parameters are estimated from a nonlinear maneuvering model before being fed into the filter. 
%
%
\subsection{Underwater Vehicles (UV)} \label{Underwater} \noindent 
Kinsey (2007) \cite{kinsey2007model, kinsey2014nonlinear} introduced an analytical development of a full state model-based observer for underwater vehicle navigation, exploiting knowledge of the UAV's nonlinear dynamics alongside Lyapunov techniques and the Kalman-Yakubovich-Popov Lemma. Morgado \cite{morgado2007vehicle, morgado2013embedded} developed a vehicle dynamics aiding technique that provides kinematic state information about the rigid body 3D motion. Hegren{\ae}s (2008) \cite{hegrenaes2007towards, Heg2008, hegrenaes2011model} proposed several model-aided implementations where not only the hydrodynamic model is addressed, but also real-time sea currents are mathematically formulated. Batista (2009) \cite{batista2009position} used ocean current estimates to calculate both velocities and position of the transponder in a closed-loop design. Lammas (2010) \cite{lammas20106} used a VD-based model to extract kinematic states from the AUV's equations of motion whenever the acoustic sensors experience outage.\\
Martinez (2015) \cite{martinez2015model} implemented a non-linear dynamic model (NLDM) for an underwater vehicle, using a 3-DoF linear dynamic model to allow online estimation of sea current parameters before and during navigation. Herlambang \cite{herlambang2015navigation} performed linearization on an AUV model using a Jacobian matrix to obtain a linear model that is then estimated by an ensemble Kalman filter (EnKF). Dinc, and later, Khaghani (2015) \cite{dinc2015integration, khaghani2016autonomous_1, khaghani2016autonomous_2}, introduced a hydrodynamic motion model of the AUV, which addresses the dynamic coupling with Coriolis and centripetal forces acting upon the vehicle. Allotta (2016) \cite{allotta2016new} introduced an innovative navigation strategy based on kinematics derived from a \textit{Typhoon} AUV hydrodynamic model and estimated by unscented KF.\\
Zhang \cite{zhang2016method} proposed a state estimation scenario around a cylindrical structure using the geometry of the special orthogonal group SO(3). Tal (2017) \cite{tal2017inertial} implemented a 6-DoF hydrodynamic model at an ELC integration scheme, simulating the AUV dynamics with respect to a corresponding rigid body motion model. Rypkema (2018) \cite{rypkema2018implementation} implemented a hydrodynamic model-based localization and navigation system where the AUV's linear velocity is determined by the measured angular rates and vehicle's propeller RPM. Arnold (2018) \cite{arnold2018robust} used a hydrodynamic-based motion model for the \textit{FlatFish} AUV to aid DVL measurements during drop outs caused by bottom locks. Peng-Fei and Karmozdi (2020) \cite{karmozdi2020implementation, lv2020underwater, karmozdi2020ins} introduced an intelligent velocity model that provides pseudo-DVL measurements obtained from kinematic states of a 3-DoF translational motion model.
\subsection{Land Vehicles (LV)} \label{land_vehicles}  \noindent 
Ma (2003) \cite{Ma2003} studied the contribution of a vehicle kinematic model (VKM) on a low-cost strapdown INS mounted on a land vehicle, in addition to body NHC. Sheu (2010) \cite{sheu2010design} introduced a hybrid of a behavior-based and a model-based scheme for robot navigation, where each module is responsible for a different level of reactive response. Salmon (2014) \cite{salmon2014exploration, salmon2015experimental} investigated how inclusion of the VDM along a closely coupled INS/GPS can improve performance of ground vehicles navigation.
Zhao (2018) and later Bijjahalli \cite{zhao2018vehicle, bijjahalli2019high}, compared performances between different operating modes of an INS/GNSS/VDM configuration, examining how vision-based sensors can contribute to the solution accuracy. Zemer (2019) \cite{zemer2019feasibility} proposed a gyro-free INS model assuming land vehicle dynamics. Thus, both linear accelerations and angular velocities are determined solely using $n$ accelerometers in a 3-DoF suitable for ground robot navigation.
\section{Conclusions} \label{sec_conc}
A comprehensive up-to-date review of information-aided navigation techniques has been carried out with the intention of providing a wide panoramic view and cover the existing literature, specifically from the last three decades. Due to its centrality, the aiding configuration was chosen to act as the classification key, concentrating the contents in three thematic categories: (i) Direct information aiding (ii) Indirect information aiding (iii) Model-based aiding. \\
Each aiding method is modelled and demonstrated with notable works, followed by analyses of the characteristic advantages and disadvantages of each use case.
In the direct information aiding section, we also provided the measurement model, which may act as a recipe for the inclusion of information aiding approaches in existing or new navigation filters.\\
As discussed at length in our paper, the INS solution is prone to drift from many directions, from instrumental noise to environmental disturbances, and its prevention is of utmost priority to the system designer. By accurately identifying the engineering challenge, one is able to choose a suitable complementary method that intelligently compensates whatsoever obstacle the system may encounter. In addition, this set of information aiding approaches requires only software modifications. 

\bibliographystyle{IEEEtran}
\bibliography{IEEEabrv, Bibliography}

\end{document}